%% file: main.tex
\documentclass[letterpaper]{article}
\usepackage{iccc}

\input{includes}
\input{macro}

\begin{document} 
\maketitle


\input{figures/summary}

\begin{abstract}
Computational creativity has contributed heavily to abstract art in modern era,
allowing artists to create high quality, abstract two dimension (2D) arts with a high level of controllability and expressibility.
However, even with computational approaches that have promising result in making concrete 3D art, computationally addressing abstract 3D art with high-quality and controllability remains an open question.
To fill this gap, we propose to explore computational creativity in making abstract 3D art by bridging evolution strategies (ES) and 3D rendering through customizable parameterization of scenes.
We demonstrate that our approach is capable of placing semi-transparent triangles in 3D scenes that, when viewed from specified angles, render into films that look like artists' specification expressed in natural language. 
This provides a new way for the artist to easily express creativity ideas for abstract 3D art.
The supplementary material, which contains \underline{code}, \underline{animation} for all figures, and \underline{more examples}, is here: \url{https://es3dart.github.io/}.
\end{abstract}

\section{Introduction}

Through art history, a trend of abstract art has been influential since the beginning of 20-th century in the course of modernism~\cite{kuiper2021modernism} which focuses on abstract elements instead of traditional photo-realistic forms.
Starting from Cubism art movement~\cite{rewald2014heilbrunn} and geometric abstraction~\cite{dabrowski2004geometric}, the focus on abstraction leads to abstract expressionism~\cite{paul2004abstract} and minimalist art~\cite{tate_minimalism,bertoni2002minimalist}. 
They collectively have opened a new approach of painting art where the subjective appreciation of the object or the feeling could be expressed, and the once dominant traditional focus on accurate representation is not the only standard anymore.

\enlargethispage{-22.0\baselineskip}

Computer art, in the broader sense making art in a computational way, have also played a heavy role in this course of abstract art. 
Early works have brought forward the concept of artists generating art by designing mathematically, or more precisely algorithmically~\cite{malkevitch2003mathematics,verostko1994algorithmic}.
The said algorithm, and its properties like its complexity~\cite{kolmogorov1965three}, 
have became an important and intrinsic metric of art~\cite{schmidhuber1997low}.
In this regard, a large body of pioneer artists including Frieder Nake, Vera Mol\'{a}r, A.\ Michael Noll,  Manfred Mohr, Leslie Mezei and Georg Nees have explored designed algorithms to produce abstract art that are composite of simple primitives like lines and polygons.
More recently, modern approaches propose that artists could, instead of designing the algorithm directly, specify rules to find an algorithm that in turn produces the artwork.
Doing so becomes feasible thanks to the recent advances in evolution strategies (ES), for example art generation using ES has been proposed to produce a wide range of  simple~\cite{johansson2008genetic,alteredqualia2008evolutiongenetic} and complex~\cite{fogleman2016,cason2016,paauw2019paintings,shahrabi2020,tian2022modern} art forms,
where the artists can specify the rules using text or images as instructions.

While arguably painting has always been one of the most dominant art forms, arts concerning three dimensional (3D) objects is an equally important field.
For example, Among 3D arts the one with the longest tradition is sculpture~\cite{rogers2020sculpture} and architecture~\cite{gowans2022architecuture} which starts from classical antiquity and remains pretty relevant today.
Yet modern techniques and industries add new movements to 3D art, where we also see a similar trend of abstractness and modernism like the painting arts mentioned above.
For example, the trend of modernism has led to the sculpture to go beyond the realm of solid, representational form, and the artists started to produce ``nonfunctional, nonrepresentational, three-dimensional works of art''~\cite{rogers2020sculpture}.
This particularly includes spatial sculpture~\cite{conry1977spatial,kricke1976spatial,caro1962early} where space becomes the subject of the 3D artworks, and the viewing angle as well as the relation of objects comes to be an important part of the art.
In the realm of computational approach to 3D art, early work explores rule based generation~\cite{broughton1997use,coates1999exploring} where the combination of rules are evaluated by human-in-the-loop~\cite{cook2007gauguin}. Late works focus on parameterization, such parametric 3D surfaces~\cite{chu2021evolving}.
A recent work~\cite{hsiao2018multi} produces wire art that looks like predefined sketches by connecting vortexes using a 3D path finding algorithm.

However, the modern computational approaches to the abstract 3D art remains an open gap to fill. 
This is more prominent given how high-quality and controllable computer \emph{concrete} 3D art has been achieved by recent advances.
For example, it is now possible to generate high-quality 3D volumetric objects using recent generative model like NeRF~\cite{mildenhall2021nerf,martin2021nerf} and text-to-image model like DALLE~\cite{ramesh2022hierarchical}.
Powerful image generative models like Imagen~\cite{saharia2022photorealistic} and Parti~\cite{yu2022scaling} open the door to works such as DreamFields~\cite{jain2022zero}, DrameFusion~\cite{poole2022dreamfusion} and Magic 3D~\cite{lin2022magic3d} where artists can easily control the generation of height quality 3D object  by text prompt.
On the other side, to our best knowledge, still missing are computational tools to produce high quality \emph{abstract} 3D art creation that does not not require artists detailing everything but instead allow artists specifying instructions in a way that is high level and that human can easily produces and understand.

\input{figures/summary-examples}

To bridge this gap, we propose to combine evolution strategies (ES) and 3D rendering through customized parameterization of scenes, which is later evaluated by a deep learning model, to address computational creativity in the abstract 3D art.
In doing so we leverage the recent advances in evolution algorithm applied to abstract 2D art generation, as well as ray-tracing rendering, which is vital to the rendering of physically-sounding transparent objects.
Two components are bridged by immediate mode, a paradigm in computer graphics where senses are parameterized.
Parameterization could be specified by the artist to customize scenes, allowing a new way for the artist to express creative ideas at a high level.
We demonstrate that our approach is capable of placing shapes in 3D scenes that, when viewing from specified angles, look like artists' specification expressed in natural language. 
This is facilitated by recent advances in deep learning, namely CLIP that is also used in DALL-E~\cite{ramesh2022hierarchical}, that connects text and images domains.
With all these components, the artist can freely express the idea of 3D abstract art by text, which is a more approachable way and allows a wider audience to participate in 3D art creativity.
A quick summary of our proposed method and some exemplary artifacts are shown in Figure~\ref{fig:summary} and Figure~\ref{fig:summary-examples}.

\input{figures/architecture}

\section{Related Works}
\label{sec:related-works}

In this section we cover works that are the background of or related to our proposed methods.

\subsubsection*{Computational Approach to Abstract Painting Art}

The computational approach to abstract and minimalist painting art has a long history before the era of computing.
Early works discuss mathematical art~\cite{malkevitch2003mathematics} which establishes the connection between artworks and mathematical properties such as symmetry and polygon for paintings, and octave for music.
Since the inception of computers as a new means for human activity,
algorithmic art~\cite{verostko1994algorithmic} has been proposed as a new framing of art, where artworks are not produced by humans directly but by human designing a mathematical process, or an algorithm, that produces the artifact.
Furthermore, the properties of the said algorithm themselves could also be a subject of artistic discussion.
One example is low-complexity art~\cite{schmidhuber1997low} where the complexity of the said algorithm becomes a measure of the artwork.
In this regard, a large body of pioneer artists have practically explored designing algorithms to produce computational abstract art that are composite of simple lines and polygons.
This includes Frieder Nake~\cite{frieder_nake},  Vera Mol\'{a}r~\cite{vera_molar}, Leslie Mezei~\cite{leslie_mezei}, A.\ Michael Noll,  Manfred Mohr and Georg Nees. 
Collectively, they represent the artists putting the early concept of computer abstract art in practice.

Naturally, artists explores whether it's possible to, instead of directly designing an algorithm, use the rules that control the possible search space of the algorithm that actually makes the art.
However, since the algorithm is hardly differentiable, gradients are not available or hard to define.
From an optimization point of view, this non-differentiability makes it challenging to find an algorithm since a wide range of optimization methods are gradient-based.
This resonates with the challenge of looking for a better neural network architecture~\cite{elsken2019neural}.
To tackle it, previous works have explored leveraging evolution strategies (ES) in art generation, since ES belongs to the category of black box optimization which does not require differentiation.
Such effort could handles art forms ranging from simple ones ~\cite{johansson2008genetic,alteredqualia2008evolutiongenetic} to more complex ones~\cite{fogleman2016,cason2016,paauw2019paintings,shahrabi2020,tian2022modern}.

\input{figures/different-number-of-triangles}

\subsubsection*{3D Rendering}
The development of Computer Graphics~\cite{foley1994introduction,shirley2009fundamentals} is largely associated~\cite{watt19933d} with the constant quest for better three dimension (3D) rendering.
One of the drives in 3D rendering is the development of game~\cite{gregory2018game} which naturally calls for high quality rendering in real-time~\cite{akenine2019real}.

Regarding rendering technique, broadly speaking two ways exist: first is rasterization~\cite{shirley2009fundamentals}, where polygons representing 3D objects are projected to pixels on 2D screen. 
It is fast, widely adopted, and often good enough. Another is ray-tracing~\cite{glassner1989introduction,spencer1962general,appel1968some,whitted2005improved}, where rays are traced back from camera, interacting with the objects it encounters accordingly to the rendering equation~\cite{kajiya1986rendering}, all the way till the light source.
It enables a high degree of physical plausibility, but at the cost of high computational requirements.

In the practice of 3D rendering engines, two paradigms exist:
One is retained mode graphics~\cite{jin2006retained} where the application issues since to graphic libraries.
This is the dominating practice due to its efficiency.
Another one is immediate mode paradigm~\cite{radich2019retained} where the application builds the scene and only issues drawing primitives to the graphic libraries. 
It is less efficient, but allows more flexibility and expression, which could helpful in creativity settings.

\subsubsection*{Evolution Strategies (ES)}
~\cite{beyer2001theory,beyer2002evolution}, as an optimization method, has been applied to many problems. 
Inspired by biological evolution, its high-level idea consists of iteratively changing parameters and keeping the sets of parameters that are most fitting. At the end of evolution the best, or the most fitting solutions remain.
A straightforward realization of this idea is iteratively  perturbing parameters randomly and keeping ones only if the change leads to better fitness.
Unfortunately, it is often computationally inefficient.
Recent advances in ES have largely improved the efficiency.
For example, PGPE~\cite{sehnke2010parameter} proposes to estimate the gradients in linear time which can be used by gradient-based optimizers like Adam~\cite{kingma2014adam} and ClipUp~\cite{toklu2020clipup}.
On the other hand, CMA-ES~\cite{hansen2000invariance,hansen2006cma} estimates the covariance matrix of parameters, which provides better performance using quadratic running time.

Notably, unlike gradient-based optimization, evolution strategies do not require the optimized problem to be differentiable, thus it could effectively serve as black-box optimization solver where only the evaluation of fitness is needed. 
This leads to a wide range of applications. 
For example, recent advances in neural evolution~\cite{such2017deep} allows efficient optimization of neural networks, and EvoJAX~\cite{tang2022evojax} fully leverages the hardware acceleration for a wide range of evolution tasks.

\input{figures/different-runs}

\subsubsection*{3D Generative Models and Computational Creativity}
One early way of generating creative 3D objects starts with 3D point cloud~\cite{nguyen20133d,guo2020deep}, which consists of points with unit volume in 3D space.
The 3D point cloud is easier to model, and is used in turn to generate the 3D shape by morphing the points~\cite{mo2019structurenet,li2021sp}.
Recently, we have seen a surge of high quality generative models that directly models 3D objects.
Especially in producing concrete, \emph{volumetric} 3D objects,
works in the line of NeRF~\cite{mildenhall2021nerf,martin2021nerf} represents the whole scene by a radiance field parameterized by the neural base models.

Research in 3D generation is not limited to the  modality of 3D objects only.
Multi-modal, text-to-image works such as DALLE~\cite{ramesh2022hierarchical}, Imagen~\cite{saharia2022photorealistic} and Parti~\cite{yu2022scaling}
allow creating high quality images using text prompts as guidance.
Based on them, text-to-3D objects have become possible.
For example, DreamFields~\cite{jain2022zero}, DrameFusion~\cite{poole2022dreamfusion}, Magic 3D~\cite{lin2022magic3d} and Imagen Video~\cite{ho2022imagen} are capably for generating photo-realistic \emph{volumetric} 3D objects following the description given in text.

Beside generative models that model concrete and real-world 3D objects, similar problems have also been approached from a computational creativity point of view, which emphasize the artistic creativity of the generated object.
Early work explores rule based generation~\cite{broughton1997use,coates1999exploring} where the combination of rules is evaluated by either enabling human-in-the-loop~\cite{cook2007gauguin} or parameterizing a single formula~\cite{chu2021evolving}.
Also, a recent work produces wire art~\cite{hsiao2018multi} resembling given sketches by first generating vortexes and then connecting them by leveraging 3D path finding algorithms.
They are probably closest to our work, but crucial differences exist:
As far as we know, we are the first work to address \emph{spatial}, \emph{abstract} 3D generation with the expressionism from modern neural based models.

\input{figures/different-transparencies}

\section{Methodology}
\label{sec:methodology}

We show the overall architecture of our proposed method in Figure~\ref{fig:architecture}.
It contains two parts, the outer loop of evolution strategies and the inner evaluation of 3D scene's fitness.

\subsection*{The outer loop of evolution strategies (ES)} This is a black box optimization that suggests multiple sets of parameters and adjusts them based on the fitness, or how well each set of parameters are.
At the end of several steps of optimization, ES gives parameters leading to better fitness.
We use CMA-ES~\cite{hansen2000invariance,hansen2006cma}, an algorithm that estimates the covariance matrix of parameters, since it provides better performance than common alternatives like PGPE~\cite{sehnke2010parameter} while only incurring marginal increase of running time in our case. 
Engineering-wise, we use EvoJAX implementation of CMA-ES, which is based on JAX~\cite{jax2018github} and runs easily on accelerators like GPUs.

\subsection*{The inner evaluation of 3D scene's fitness}
In our setting, the parameters literally parameterize the building and the rendering of 3D scenes.
We first build the 3D scene, and then render it from multiple, user-specified cameras using a ray-tracing renderer engine.
While the actual spatial objects are parameterized by the parameters, how the builder and the renderer interpret these parameters are considered hyper-parameters that the artist users could control.
Finally, the rendered images from each camera, or ``films'' following photography terms, are compared with the text prompts semantically, which is done by computing the Cosine loss between images and texts encoded by CLIP encoders.
The mean loss of all pairs of images and texts is given back to the aforementioned evolution strategy for adjusting the parameters accordingly.

Overall, the evaluation of 3D scenes through parameterization is the main contribution we device to help artists express creativity.
We detail the key decisions as follows:

\subsubsection*{Parameterization}
Since our goal focuses on the computational creativity of \emph{spatial} 3D art as motivated by the trend of abstract art in modern sculpture, we choose placing semi-transparent triangles in plastic material in the 3D space.
Concretely, each of $N$ triangles is associated with $13$ learnable parameters,
namely the position of its three vertices $(x_1,y_1,z_1)$, $(x_2,y_2,z_2)$, $(x_3,y_3,z_3)$ and the color and transparency $(R,G,B,A)$, thus making totally $13N$ parameters.
It is possible to archive photo-realistic rendering of these semi-transparent triangles with the help of a ray-tracing renderer, because light rays may pass through and bounce between them many times.
Furthermore, doing so allowing retain the possibility of reproducing a solution in the real world.
The choice of semi-transparent triangles is inspired by a recent work on 2D abstract art~\cite{tian2022modern}, but going to 3D, as in our setting, makes our whole new pipeline necessary since the technique and the optimization dynamics are completely different.

\subsubsection*{Rendering}
In practice, for ray-tracing rendering we use physic-based Mitsuba 3 renderer~\cite{nimier2019mitsuba}.
For each triangle, the bidirectional scattering distribution function (BSDF) for rendering is set as a mixture of BK7 Glass, a thin dielectric material, and a \emph{Lambertian}, an ideally diffuse material of corresponding $R,G,B$ value, mixed with ratio $A$.
Besides physically correct rendering, Mitsuba 3 also allows GPU-powered paralleling sampling, which largely accelerates the ray-tracing rendering.
For the sake of completeness, we note that Mitsuba 3 is also capable of sampling-based gradient estimation, but we do not leverage such capacity and leave study of that behavior as an orthogonal research direction for future study.

\input{figures/different-prompt-texts}

\subsubsection*{Evaluation}
As we expect the pipeline to produce a scene that, when rendered from different cameras (we call what a camera produces ``films''), looks like corresponding text semantically,
We measure it using CLIP, which provides an image and a text encoder that projects images and text into a shared, comparable latent space with Cosine distance.
Note that with multiple pairs of camera and text, we could make the produced 3D object look like (or different) from different directions.
In doing so, each film is encoded and compared with corresponding encoded text, and the means of Cosine distance of all such pairs are used as fitness, which is given back to the evolution strategies.

\subsection*{Computation Platform}
Since both the rendering and the evolution strategies we use are fully run-able on GPU, the computation is fast and in our experience can be tens of times faster than on CPU, thus fully leveraging the modern hardware accelerators.

\section{Experiments}
\label{sec:experiments}

In this section we showcase our method with several experiments.
In Figure~\ref{fig:summary-examples}, we show several examples of the evolved 3D art produced by our method, each with $1,200$ steps of evolution and a population of $128$ using CMA-ES.
As shown here, our method demonstrates that a wide range of text prompts can be handled by our method, producing spatial, abstract art that is both novel and consistent with human interpretation.
Even given the abstract nature defined by the scene, our method could still handle both the spatial shape (first two examples) and the color (last two examples).

In the rest of this section, we investigate how turning several important hyper-parameters could impact the finally generated 3D art, showing the dynamics of our method which could be served as a guidance for the artist users.

\subsubsection*{Different Number of Triangles}
The number of triangles could be used as a kind of ``budget'' that our method used to allocate in occupying the 3D space. 
In Figure~\ref{fig:different-number-of-triangles}, we show our method generating 3D artwork with 10, 25, 50 and 100 triangles, respectively.
Our method thus places triangles in the increasing order of granularity,
where the general outline is first emphasized and details are then filled.
Note that the number of parameters increases in proportion to the number of triangles, requiring more computation time in rendering and optimization.
Thus, this is a balance that artist users should decide.

\subsubsection*{Different Runs of the Same Configuration}
One important aspect of computational creativity is the ability to produce variants of the art given the same instruction. 
Such a property not only allows artists to be ``in-the-loop'' to choose from a wide range of variants, but also shows the capacity of the generative model.
In Figure~\ref{fig:different-runs}, we show two configurations, each with two independent runs.
It is shown that different runs lead to equally plausible yet largely different 3D art.
In doing so, our method could support artists ``in-the-loop'' of the creativity process.
    
\subsubsection*{Fixed v.s. Learnable Transparency}

Unlike 2D art, in 3D art the transparency matters a lot, due to the effects such as reflection and optical diffusion. 
This is especially true in the spatial setting as we focus on.
In Figure~\ref{fig:different-transparencies}, we demonstrate several settings of transparency, including the fixed transparency and the default setting of learnable transparency.
It shows that while fixed transparency allows a more consistent global outlook, it nonetheless limits the expression by forcing large and small triangles contributing the same to the images. 
In contrast, learnable transparency gives our method flexibility in how the triangles are related to the 3D space.

\subsubsection*{Different Text Prompts at Different Cameras}
While in many examples we show the same text prompt for cameras, this is completely not a requirement imposed by our method.
On the contrary, our method allows pairs of texts and cameras in an arbitrary combination.
Such a capacity allows a wider range of creativity from users to, for example, generate a 3D art that looks differently from different angles.
In Figure~\ref{fig:different-prompt-texts} we demonstrate one such case,
where our method generates 3D art that looks like ``Walt Disney World'' from two directions but ``an annoyed cat'' from the other two directions,
even these views are of the same, single 3D art.
We argue that our method is the first to be capable of helping artists in such a creative process that previously requires lots of manual work~\cite{hsiao2018multi}.

\section{Conclusion}
\label{sec:conclusion}

In this work we address the problem that is previously not studied --- generating 3D, abstract, and spatial art that is semantically aligned with human interpretation.
In doing so, we propose to leverage evolution strategies (ES) with ray-tracing rendering of parameterized 3D scenes, along with CLIP method for measuring the semantic similarity.
We demonstrate that our approach is capable of producing 3D arts through several experiments, and provides the flexibility for artists or users to fine-tune for the desired result.

Nonetheless, our proposed method is best suited as a call for further future study in computational approaches in 3D art.
For example, it remains unclear whether optimization using differentiability of the renderer would lead to a different dynamic and thus art style.
Also, the designing of the parameterized scene is a time-consuming one requiring the extensive knowledge of 3D rendering and optimization, so whether it would be improved through the (semi-)automation process should be studied too.

\clearpage
\bibliographystyle{iccc}
\bibliography{references}

\end{document}

%% file: includes.tex
\usepackage[utf8]{inputenc} 
\usepackage[T1]{fontenc}    

\usepackage{url}            
\usepackage{booktabs}       
\usepackage{amsfonts}       
\usepackage{nicefrac}       
\usepackage{microtype}      
\usepackage[table,xcdraw,dvipsnames]{xcolor} 
\usepackage{graphicx}
\usepackage[labelfont={bf}]{caption}
\usepackage{subcaption}
\usepackage{footnote}
\usepackage{booktabs}
\usepackage{algorithm}
\usepackage{algorithmic}
\usepackage{wrapfig}
\usepackage{enumitem}
\usepackage{amssymb,amsmath}
\usepackage{mathtools}
\usepackage{multirow}
\usepackage[normalem]{ulem} 
\usepackage{comment}
\usepackage{longtable}
\usepackage{todonotes}
\usepackage{setspace}
\usepackage{comment}

\usepackage{times}
\usepackage{helvet}
\usepackage{courier}

\graphicspath{ {./images/} }

%% file: macro.tex
\title{Evolving Three Dimension (3D) Abstract Art: \\Fitting Concepts by Language}

\author{Yingtao Tian\\
Google Research, Brain Team\\
\texttt{alantian@google.com}\\
}

%% file: figures/summary.tex

\begin{figure}[!b]\setlength{\hfuzz}{1.1\columnwidth}
\begin{minipage}{\textwidth}
    \centering
    \includegraphics[trim=0 180px 0 0, clip, width=1.0\textwidth]{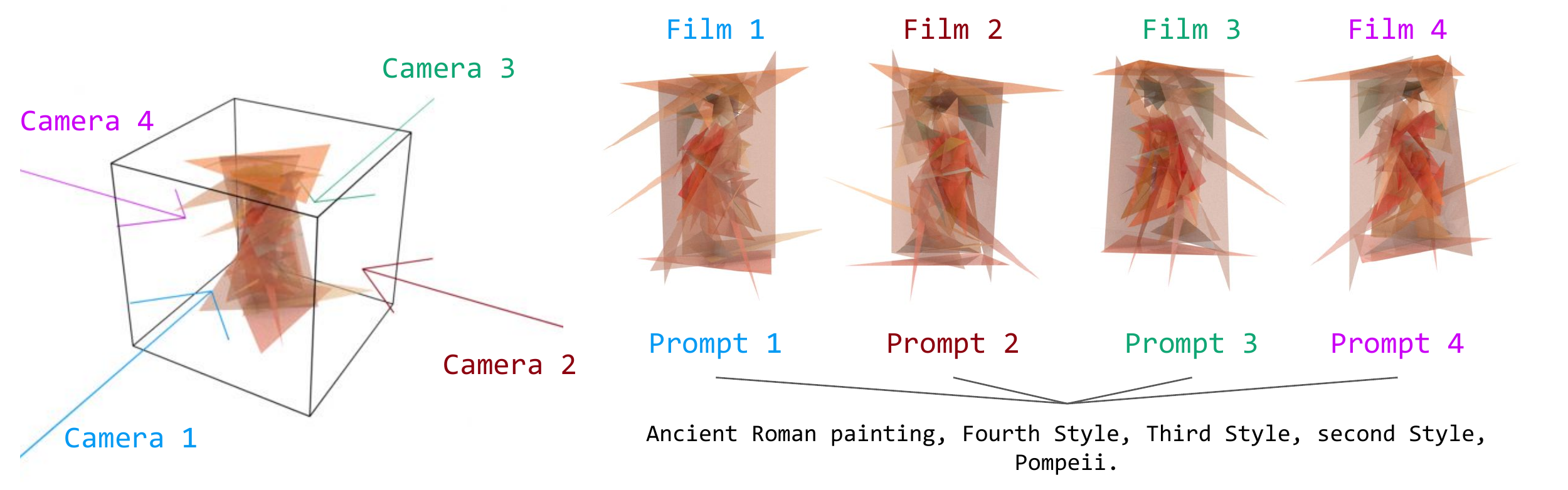}
    \caption{
    Our proposed method places semi-transparent triangles in three dimension (3D) spaces using Evolution Strategies~\protect\cite{tang2022evojax,hansen2000invariance,hansen2006cma}.
    Leveraging ray-tracing based rendering Mitsuba 3~\protect\cite{nimier2019mitsuba,Jakob2020DrJit}, the rendered film at possibly multiple cameras is compared with its corresponding, user-specified text prompt using distance between their representation embedded by CLIP~\protect\cite{radford2021learning}.
    Such distances, aggregate by average, are used as the fitness in the sense of Evolution Strategies, which optimize the parameters of triangles to achieve better finesses.
    }
    \label{fig:summary}
\end{minipage}
\end{figure}

%% file: figures/summary-examples.tex
\begin{figure*}[!htbp]

    \centering

    \begin{subfigure}{1.0\textwidth}
        \centering
        \includegraphics[page=1, trim=0 230px 0 0, clip, width=1.0\textwidth]{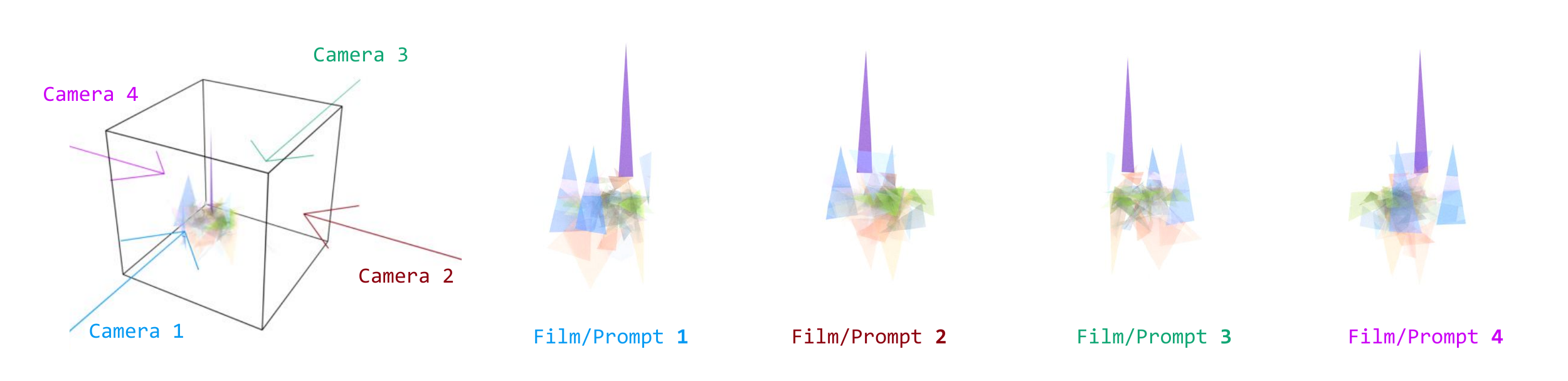}
        \caption{The prompt for all camera/films is ``Walt Disney World''}
    \end{subfigure}

    \begin{subfigure}{1.0\textwidth}
        \centering
        \includegraphics[page=2, trim=0 230px 0 0, clip, width=1.0\textwidth]{images/figures-summary-examples/figures-summary-examples.pdf}
        \caption{The prompt for all camera/films is ``A painting of Human''}
    \end{subfigure}

    \begin{subfigure}{1.0\textwidth}
        \centering
        \includegraphics[page=3, trim=0 230px 0 0, clip, width=1.0\textwidth]{images/figures-summary-examples/figures-summary-examples.pdf}
        \caption{The prompt for all camera/films is ``A bright, vibrant, dynamic, spirited, vivid painting of a dog.''}
    \end{subfigure}
    
    \begin{subfigure}{1.0\textwidth}
        \centering
        \includegraphics[page=4, trim=0 230px 0 0, clip, width=1.0\textwidth]{images/figures-summary-examples/figures-summary-examples.pdf}
        \caption{The prompt for all camera/films is ``A vivid, colorful bird''}
    \end{subfigure}

    \caption{
    Several examples of the abstract 3D art produced by our method, where the evolution process places triangles inside the unit cube space (visualized by black frame) and sets triangles' colors and transparencies, forming a spatial configuration.
    In each example shown here, four cameras look at the unit cube space from four sides, although this is an arbitrary decision and cameras can have different numbers and directions.
    The film from each camera, capturing the rendered images, is compared with the prompt.
    It could be observed that our method is capable of making a 3D art, which follows the spatial abstract art style, that looks like what humans can compose in natural language text.}
    
    \label{fig:summary-examples}
    
\end{figure*}

%% file: figures/architecture.tex
\begin{figure}[!htbp]

    \centering
    \captionsetup[subfigure]{labelformat=empty}

    \includegraphics[trim=0 170px 0 0, clip, width=1.0\columnwidth]{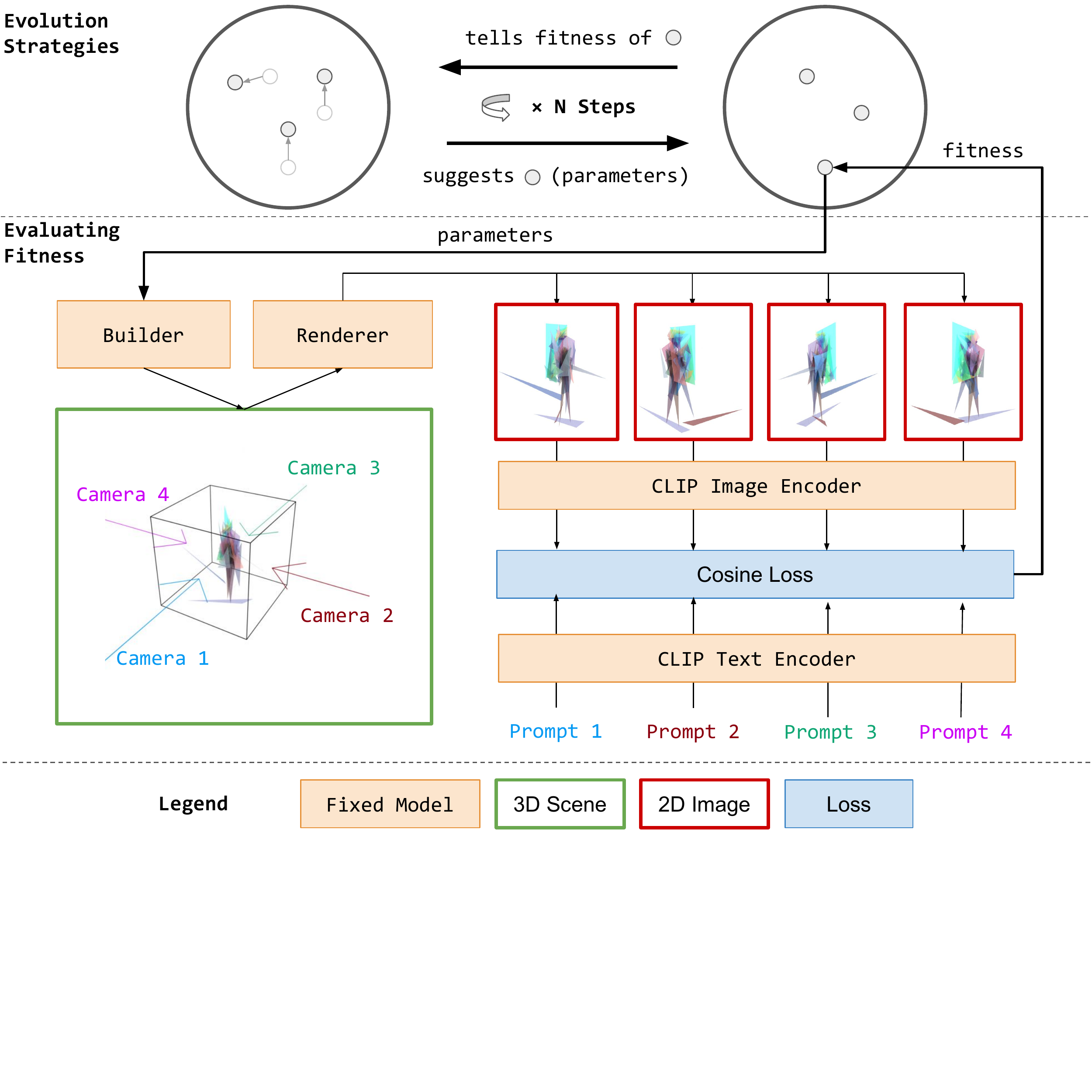}
        
    \caption{The architecture of our method, consisting of an outer loop of evolution strategies (ES) to find parameters leading to better fitness, and an inner, actual evaluation of fitness.
    The builder builds spatial 3D objects that compose semi-transparent triangles on the 3D space from parameters.
    The renderer renders the 3D space from different cameras producing corresponding images or  ``films'' which are compared with provided text prompts using Cosine loss between the images and text prompts encoded by CLIP encoders.
    Such a loss is treated as the fitness of the parameters given back to ES.
    The user of our proposed method specifies the text prompt and hyper-parameters governing the behavior of the builder and the renderer, allowing expressing creativity.
    }

    \label{fig:architecture}
    
\end{figure}

%% file: figures/different-number-of-triangles.tex
\begin{figure*}[!htbp]

    \centering

    \begin{subfigure}{1.0\textwidth}
        \centering
        \includegraphics[page=1, trim=0 230px 0 0, clip, width=1.0\textwidth]{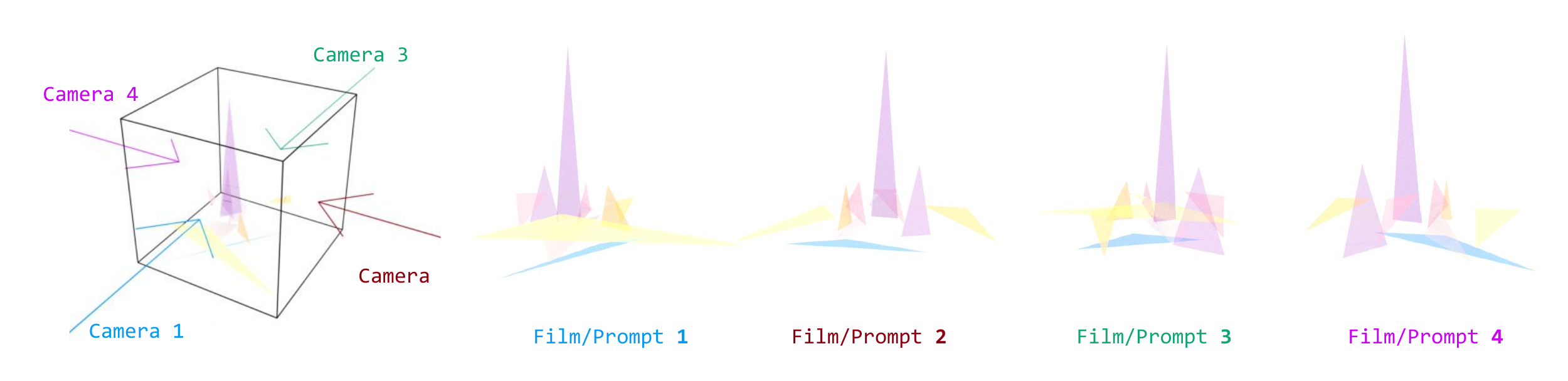}
        \vskip -5px
        \caption{``Walt Disney World'' - $10$ Triangles.}
    \end{subfigure}

    \begin{subfigure}{1.0\textwidth}
        \centering
        \includegraphics[page=2, trim=0 230px 0 0, clip, width=1.0\textwidth]{images/figures-different-number-of-triangles/figures-different-number-of-triangles.pdf}
        \vskip -5px
        \caption{``Walt Disney World'' - $25$ Triangles.}
    \end{subfigure}

    \begin{subfigure}{1.0\textwidth}
        \centering
        \includegraphics[page=3, trim=0 230px 0 0, clip, width=1.0\textwidth]{images/figures-different-number-of-triangles/figures-different-number-of-triangles.pdf}
        \vskip -5px
        \caption{``Walt Disney World'' - $50$ Triangles.}
    \end{subfigure}
    
    \begin{subfigure}{1.0\textwidth}
        \centering
        \includegraphics[page=4, trim=0 230px 0 0, clip, width=1.0\textwidth]{images/figures-different-number-of-triangles/figures-different-number-of-triangles.pdf}
        \vskip -5px
        \caption{``Walt Disney World'' - $100$ Triangles.}
    \end{subfigure}

    \caption{Our method generating with text prompts ``Walt Disney World'' with four cameras, 
    with different numbers of triangles, namely 10, 25, 50 and 100 respectively.
    It could be shown that our method leverages the budgets of triangles in the increasing order of granularity, by first using triangles for general shape and then moving towards fine-grained details. }
    
    \label{fig:different-number-of-triangles}

\end{figure*}

%% file: figures/different-runs.tex
\begin{figure*}[!htbp]

    \centering

    \begin{subfigure}{1.0\textwidth}
        \centering
        \includegraphics[page=3, trim=0 230px 0 0, clip, width=1.0\textwidth]{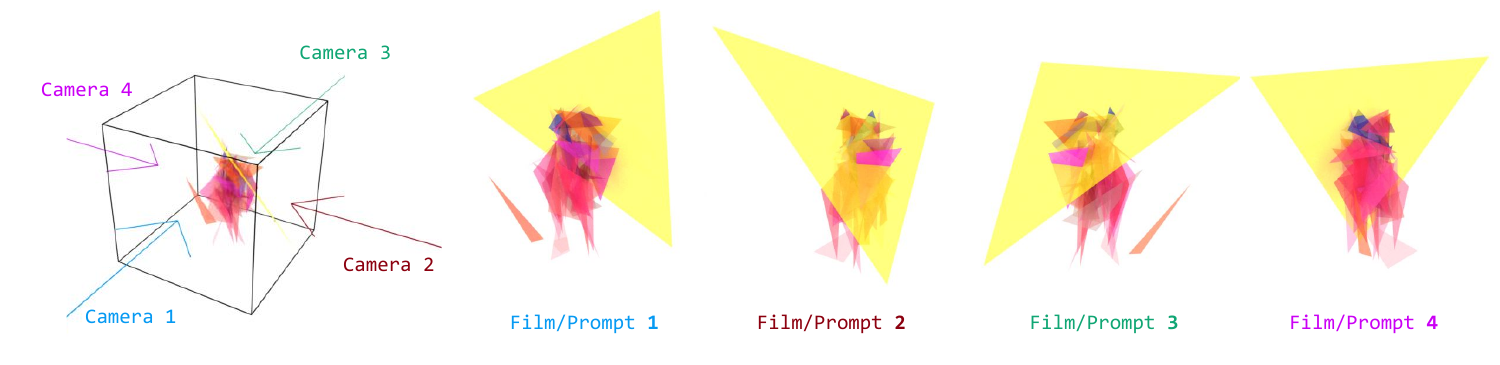}
        \vskip -2px
        \caption{``A bright, vibrant, dynamic, spirited, vivid painting of a dog'' - Run 1}
    \end{subfigure}
    
    \begin{subfigure}{1.0\textwidth}
        \centering
        \includegraphics[page=4, trim=0 230px 0 0, clip, width=1.0\textwidth]{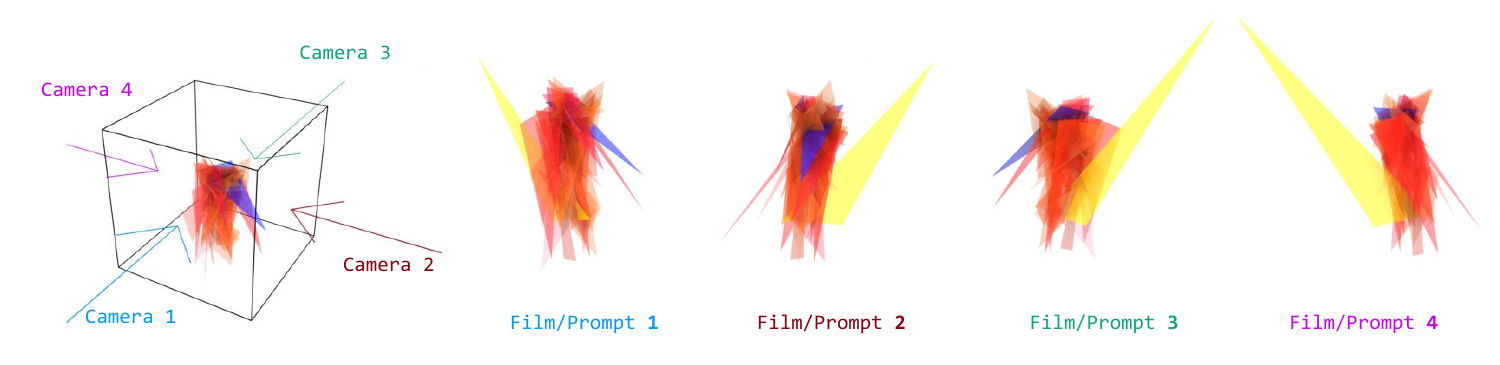}
        \vskip -2px
        \caption{``A bright, vibrant, dynamic, spirited, vivid painting of a dog'' - Run 2}
    \end{subfigure}

    \caption{Our method generates with two independent runs, both with prompt ``A bright, vibrant, dynamic, spirited, vivid painting of a dog'' from four directions.
    Different runs lead to equally plausible yet largely different 3D art.
    An artist user could exercise discretion ``in-the-loop'' by choosing from different variants from these runs.
    }
    
    \label{fig:different-runs}
\end{figure*}

%% file: figures/different-transparencies.tex
\begin{figure*}[!htbp]

    \centering

    \begin{subfigure}{1.0\textwidth}
        \centering
        \includegraphics[page=2, trim=0 230px 0 0, clip, width=1.0\textwidth]{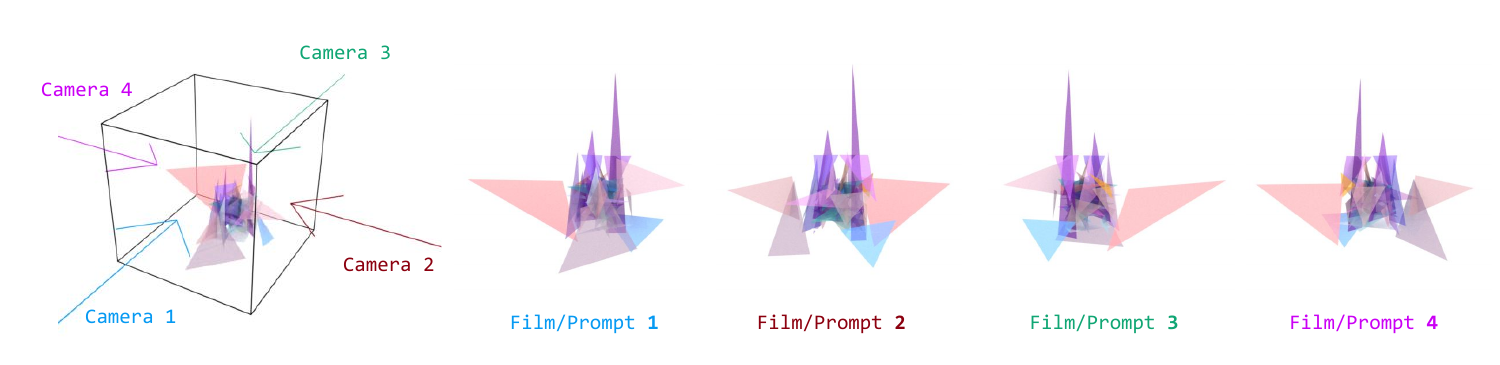}
        \vskip -4px
        \caption{Fixed transparency of $50\%$.}
    \end{subfigure}

    \begin{subfigure}{1.0\textwidth}
        \centering
        \includegraphics[page=4, trim=0 230px 0 0, clip, width=1.0\textwidth]{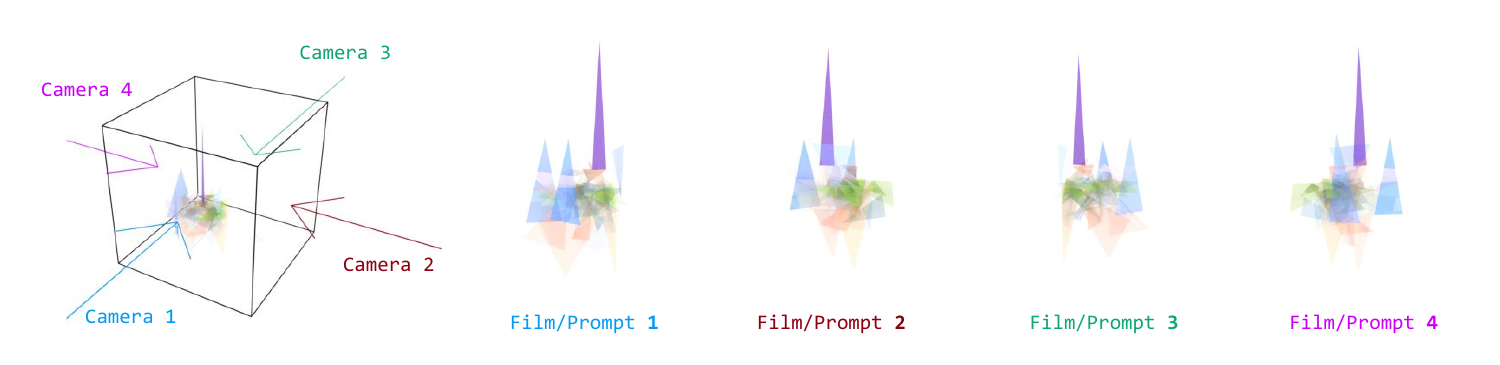}
        \vskip -4px
        \caption{Learnable transparency.}
    \end{subfigure}

    \caption{Our method generates with text prompts ``Walt Disney World'', with a fixed transparency of $50\%$ and with the default setting of learnable transparency.
    While the fixed transparency setting allows more global control of the scene, the learnable one provides great flexibility in how triangles are related to the space.
    More results from different transparency ($0\%$ and $80\%$) could be found in supplementary materials online for comparison.
    }
    
    \label{fig:different-transparencies}
    
\end{figure*}

%% file: figures/different-prompt-texts.tex
\begin{figure*}[!hbtp]

    \centering
    \captionsetup[subfigure]{labelformat=empty}

    \vskip 5px
    \includegraphics[page=1, trim=0 180px 0 0, clip, width=1.0\textwidth]{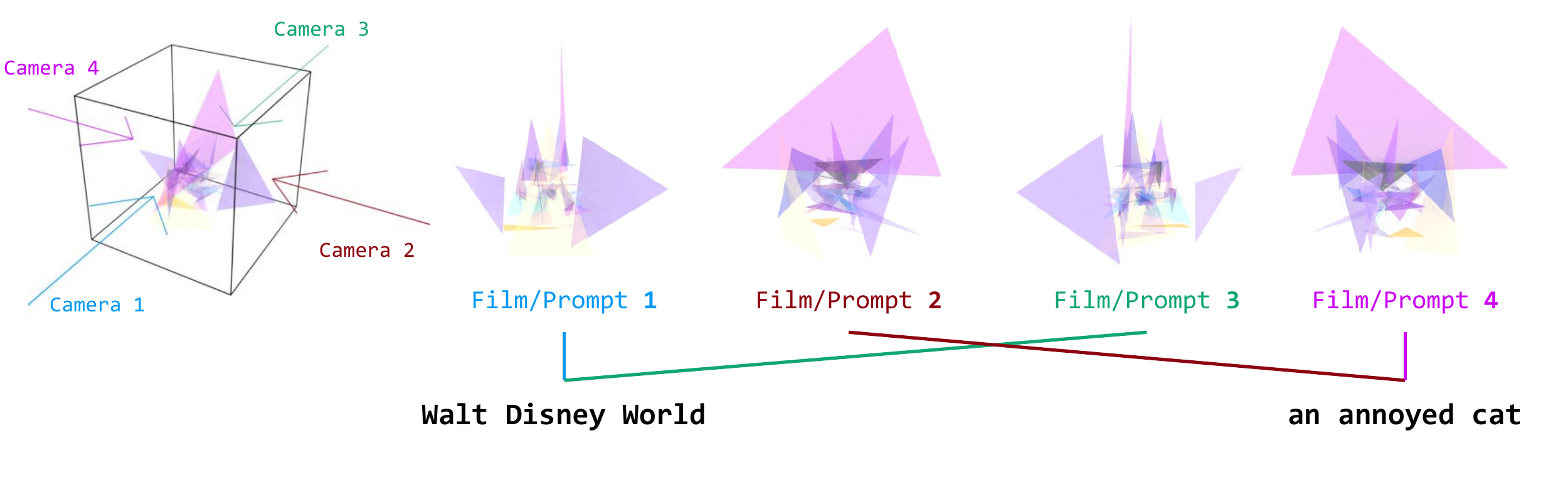}
    
    \caption{Our method generates with different text prompts at cameras.
    The text prompt for camera 1 and 3 is ``Walt Disney World'' and for camera 2 and 4 is ``an annoyed cat''.
    Our method produces one 3D art, and successfully allows it to look differently from different angles.
    }

    \label{fig:different-prompt-texts}

\end{figure*}